%% file: manuscript_arXiv.tex
\documentclass[runningheads]{llncs}
\usepackage[T1]{fontenc}
\usepackage{multirow}

\usepackage{graphicx}
\usepackage[caption=false]{subfig}
\usepackage{amsmath, amssymb}
\usepackage{cleveref}
\usepackage{eso-pic} 

\AddToShipoutPictureBG*{%
	\AtPageUpperLeft{%
		\hspace*{\dimexpr0.5\paperwidth-8cm\relax} 
		\raisebox{-1cm}{\textcolor{red}{\textbf{Presented in MICCAI Workshop on Advancing Data Solutions in Medical Imaging AI 2024}}}%
	}%
}

\begin{document}

\title{Label Sharing Incremental Learning Framework for Independent Multi-Label Segmentation Tasks}
\titlerunning{Label Sharing Incremental Learning Framework}
%
\author{Deepa Anand\and
Bipul Das \and Vyshnav Dangeti \and Antony Jerald \and Rakesh Mullick \and Uday Patil \and Pakhi Sharma \and Prasad Sudhakar}
\authorrunning{Deepa Anand et al.}
\institute{Science and Technology Organization, GE HealthCare, Bangalore, India}
\maketitle              
\begin{abstract}
In a setting where segmentation models have to be built for multiple datasets, each with its own corresponding label set, a straightforward way is to learn one model for every dataset and its labels. Alternatively, multi-task architectures with shared encoders and multiple segmentation heads or shared weights with compound labels can also be made use of. This work proposes a novel “label sharing” framework where a shared common label space is constructed and each of the individual label sets are systematically mapped to the common labels. This transforms multiple datasets with disparate label sets into a single large dataset with shared labels, and therefore all the segmentation tasks can be addressed by learning a single model. This eliminates the need for task specific adaptations in network architectures and also results in parameter and data efficient models. Furthermore, label sharing framework is naturally amenable for incremental learning where segmentations for new datasets can be easily learnt. We experimentally validate our method on various medical image segmentation datasets, each involving multi-label segmentation. Furthermore, we demonstrate the efficacy of the proposed method in terms of performance and incremental learning ability vis-\`a-vis alternative methods.  
\end{abstract}
%
%
%
\section{Introduction}
\label{sec:intro}
\input{sections/intro}

\section{Methodology}
\label{sec:method}
\input{sections/method}
\section{Experiments}

\label{sec:expt}
\input{sections/expt}

\section{Discussion}
\label{sec:disc}
\input{sections/discussion}



%
%
%
%
%

\input{manuscript_arXiv.bbl}
\newpage

\begin{center}
	{\Large \textbf{Appendix}}
\end{center}

\appendix
\section{Labels}
\begin{table}[h]
	\addtolength{\tabcolsep}{-1pt}
	\centering
	\caption{Dataset Group 1 -  Anatomy segmentation - Task and label Description}
	\begin{tabular}{|c|c|c|c|c|c|}
		\hline
		\multirow{2}{*}{\textbf{Task Names}} & \multicolumn{5}{c|}{\textbf{Labels}} \\
		\cline{2-6}
		& \textbf{Label-1} & \textbf{Label-2} & \textbf{Label-3} & \textbf{Label-4} & \textbf{Label-5}  \\ \hline
		\textbf{Task1 - Pelvic region}  &  Gluteus Minimus &	Gluteus Maximus	& Hip &	Femur &	Gluteus Medius \\ \hline
		\textbf{Task 2 - Abdomen region}  &  Liver	& Kidney & Stomach & Spleen & Gallbladder \\ \hline
		\textbf{Task 3 - Chest Region}  &  Lung	& Scapula	& Humerus & Clavicula & \\ \hline
		\textbf{IL Task- Head Region} & Mandible & Partoid  & Eyes & & \\ \hline
	\end{tabular}
	\label{tab:tasks1}
\end{table}

\begin{table}[h]
	\addtolength{\tabcolsep}{-1pt}
	\centering
	\caption{Dataset Group 2 - Lower Extremities Localization - Task and label Description}
		\begin{tabular}{|p{1.5cm}|c|c|c|c|c|}
			\hline
			\multirow{2}{*}{\textbf{Task Names}} & \multicolumn{5}{c|}{\textbf{Labels}} \\
			\cline{2-6}
			& \textbf{Label-1} & \textbf{Label-2} & \textbf{Label-3} & \textbf{Label-4} & \textbf{Label-5}  \\ \hline
			\textbf{Hip Axial} & Femur Head & Iliac & Ishiac Spine & N/A & N/A  \\ \hline
			\textbf{Hip Coronal} & Femur Head & Iliac & Ishiac Spine & N/A  & N/A \\ \hline
			\textbf{Knee Axial} & Condyles & N/A & N/A & N/A & N/A \\ \hline
			\textbf{Knee Coronal} & Tibia & Plateau & Femur & N/A & N/A \\ \hline
			\textbf{Knee Sagittal} & Tibia & Plateau & Femur & N/A & N/A \\ \hline
			\textbf{Foot Axial} &  First Metatarsal &	Third Metatarsal & Talo-Navicular Joint & Medial Malleolus & Lateral Malleolus \\ \hline
			\textbf{Foot Coronal Back} &  Sub-Taloid Joint	& Tibia & N/A& N/A&N/A \\ \hline
			\textbf{Foot Coronal Front} & First Metatarsal & Second Metatarsal & Third Metatarsal & N/A& N/A\\ \hline
			\textbf{Foot Sagittal}  & First Metatarsal & Talo-Navicular Joint & Sub-Taloid Joint & Tibia & N/A  \\ \hline
			\textbf{IL - Head Axial}  & Eye & Ear & Nasian & N/A& N/A  \\ \hline	
			\textbf{IL - Head Coronal}  & Eye & Ear & Nasian & N/A&N/A   \\ \hline	
		\end{tabular}
	\label{tab:tasks2}
\end{table}

\end{document}

%% file: sections/intro.tex

Semantic segmentation in medical imaging a fundamental task and has immense significance. By precisely delineating distinct anatomical structures or pathologies within medical images (such as MRI, CT scans, or X-rays), semantic segmentation contributes to accurate diagnosis, effective treatment planning, and disease monitoring and decision making. While deep learning (DL) solutions have emerged as the de-facto standard for image segmentation tasks, data and compute requirements remain substantial due to the sheer variety and complexity of segmentation tasks that exists. 


In the context of multi-task segmentation, where models need to handle multiple multi-label segmentations across various anatomical regions, several solutions have been proposed to build models efficiently with shared parametrization of the models. 
Examples include shared encoder with task-specific decoders~\cite{maack2024efficient},~\cite{ullah2023ssmd} and universal models with individual channels~\cite{gu2024dataseg}. Though these approaches are simple and intuitive, performance degradations are observed when the number of tasks increase.

A second set of methods employ unified label spaces to handle multi-label segmentation problems on multiple (partially labelled) datasets~\cite{lambert2020mseg}. 
Authors in \cite{gu2024dataseg}  propose a multi-dataset multi-task segmentation model that leverages text embedding of semantic classes to improve performance across across panoptic, semantic, and instance segmentation tasks. A similar approach is taken in~\cite{gong2021mdalu} for datasets with partial annotations.

Other methods for universal segmentation model for multi-dataset segmentation use an additional task type input to dynamically construct filters for shaping the segmentation outputs according to the type of dataset~\cite{zhang2021dodnet}. In~\cite{huang20193d}, the authors propose a mechanism to utilise task specific filters using adapters. Most of approaches listed here assume that the data across multiple datasets have semantic similarity, albeit having disparity in the label space.

Existing methods for multi-task segmentation often exhibit inherent limitations when it comes to accommodating new tasks. These models are typically tailored to a predefined set of tasks and lack extensibility for incorporating additional ones. Moreover, they struggle with incremental learning, risking catastrophic forgetting of previously learned tasks. While various approaches address this challenge, most require architectural modifications for each new label set.

In this paper, we propose a novel plug-and-play ``label sharing'' framework that aims to address these limitations. Our approach involves learning a single model across multiple independent segmentation tasks, each associated with multiple labels. Through our experiments, we demonstrate that assigning common shared labels to disparate labels across tasks/datasets, in the simplest possible way, is sufficient for learning a model whose performance is on par with multi-task schemes. Moreover, our method enables seamless updation for new segmentation tasks without compromising overall accuracy. We evaluate the label sharing framework across diverse medical image segmentation tasks, each involving varying numbers of labels. Both qualitative and quantitative analyses demonstrate its effectiveness and versatility.

%% file: sections/method.tex

Consider $M$ medical image segmentation tasks\footnote{We use the term `task' to mean multi-label segmentation on a given dataset pertaining to a certain anatomy.} $T_1, T_2, \cdots, T_M$ with $n_1, n_2, \cdots, n_M$ labels respectively. Let $\ell_{ij}$ be the anatomy corresponding to the $j^{th}$ label of the $i^{th}$ task. The label sharing framework proposes to group the labels across different tasks and assign a shared abstract label $\mathrm{\Lambda}_k$ for each group $k$, with the following constraints:
\begin{enumerate}
\item Every label $\ell_{ij}$ gets assigned to exactly one shared label $\mathrm\Lambda_k$. 
\item In every shared label group $\mathrm\Lambda_k$, there is at most one label from each of the tasks $T_i$. 
\end{enumerate}

\begin{figure}[t]
\begin{center}
\includegraphics[width=\textwidth]{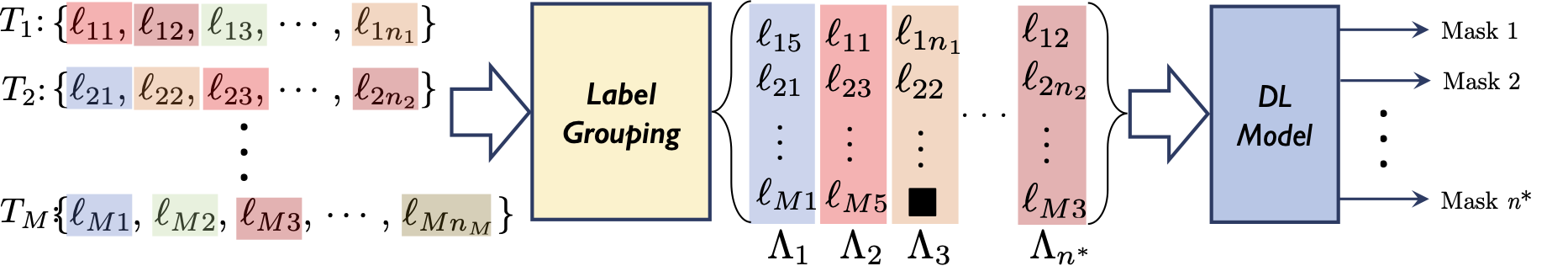}
\caption{Schematic of the proposed label sharing framework. The tasks with identically coloured boxes on the left share a common attribute and hence they are included in the same shared label, as shown after the label grouping block. The black square $\blacksquare$ represents the unavailability of a label from a particular task to be included.}
\label{fig:scheme}
\end{center}
\end{figure}

Essentially, label sharing is a partitioning of union of all the individual labels of all the tasks. The two constraints together imply that the total number of label groups $n^*$ is at least as big as the largest number of labels in individual tasks: $n^*\geq n_{\text{max}}:= \max_i n_i$.~\Cref{fig:scheme} is the schematic depiction of the proposed label sharing framework. As shown, a single common deep learning model can be trained on the combined datasets with shared labels. The second constraint also means that a few shared labels may not have representation from certain tasks and this is denoted by a solid square in the depiction.

%
%

\subsection{Label sharing}
Label sharing has to be such that the labels within each $\mathrm\Lambda_k$ share a set of common characteristics, thereby aiding training of the segmentation model. It is possible to consider several attributes of either the segmentation masks and/or the underlying images to perform grouping. However, in this work we show that it is sufficient to simply use the average relative sizes (within the tasks) of the masks to be segmented as the attribute. We sort the labels within tasks according to their relative sizes and match them across tasks, with ties broken arbitrarily. This scheme satisfies both the constraints mentioned earlier, and also results in $n^*=n_{\text{max}}$. 

Now that the labels are harmonized across tasks, we propose to build a single multi-channel neural network model to solve the segmentation problem in the shared label space. This makes the number of parameters in the last layer of the neural network grow in the order of only $\mathcal O(n_{\text{max}})$. Additionally, there is no need for any architectural or training procedure modifications to supply the network with side information about the specific task to be solved. 

\subsection{Task addition}
The label sharing framework provides a natural way to incorporate new tasks to the segmentation network in a plug-and-play fashion, with almost no computational or storage overhead. This involves only computing the average relative sizes of the structures to be segmented in the new task and mapping the individual labels to the shared labels. Therefore, this requires only storing a table of average relative sizes of the structures of all the tasks used to train the model hitherto. However, the number of labels in the new task should be less than or equal to $n^*$. 

In the next section, we present experimental results that support our claims.

%% file: sections/expt.tex
%
%
%

The performance of our proposed method was assessed across two distinct use case scenarios - (i) anatomy segmentation in a dataset of 2D image slices and (ii) labeling/localizing extremity structures in a dataset of 2D projections. The proposed label sharing approach was employed in two different modes: (a) label-sharing (LS) with all tasks considered simultaneously and (b) incremental learning (IL), where tasks were learned progressively.

We compared our proposed label sharing approach with the following three alternative approaches (i) Individual Models: a dedicated model was built for each multi-label task (ii) Merged Multi-Channel Model: All tasks were combined into a single multi-channel model, where each output channel corresponded to an output label; (iii) Network models with filters guided by task-specific controllers~\cite{zhang2021dodnet}.
Our findings shed light on the effectiveness of label sharing approach in comparison to these alternative strategies.

\subsection{Datasets} 
We used two different datasets for two use-cases in our experiments. 
\subsubsection{2D segmentation use-case}
The dataset for the segmentation use-case in 2D comprises a subset of data sourced from TotalSegmentator~\cite{wasserthal2022totalsegmentator}. We subdivided the data into three major anatomical segments, which, according to earlier definition, are the individual tasks: (1) pelvic (2) abdomen and (3) chest, with 5, 5, and 4 labels respectively as detailed in Table 2 of appendix A. To ensure visual domain disjointness for these tasks, we meticulously removed transition slices—such as those between the liver (Task 2) and thoracic regions (Task 3)—from the training images of both tasks. Our training dataset comprised $330$, $224$, and $263$ image volumes, while the testing dataset included $99$, $65$, and $66$ image volumes for Task $1$, Task $2$, and Task $3$, respectively. Additionally, we introduce the Head and Neck dataset~\cite{podobnik2023han} with 3 labels, consisting of $36$ training and $6$ test image volumes, to evaluate the incremental learning (IL Task) capability when incorporating additional tasks.

\subsubsection{Extremity localization task in 2D}

The second dataset consists of 2D orthogonal projections derived from computed tomography (CT) datasets, for annotating various lower extremity anatomical structures. We use this dataset for evaluating the proposed method for localization or detection tasks on medical images, crucial for several applications such as registration, detection of pathology, etc. The localization task is essentially also a segmentation task, but the annotations are either straight lines or small spherical regions in 3D. Our in-house dataset consists of knee, ankle, hip, and foot exams. Trained radiologists marked up to 5 important landmarks (hence up to 5 segmentation labels) on 3D images within each of these regions. Subsequently, mean intensity projected (MIP) images were generated from the corresponding orthogonal (axial, coronal, and sagittal) projected masks based on these ground truth markings. Therefore, one 3D volume gives rise to one 2D image in each projected direction. Notably, the projected images along the three projected direction exhibit distinct appearances, leading us to treat them as separate tasks. The specific tasks and associated labels are detailed in Table 3 of Appendix A. Figure ~\ref{fig:Extrem} shows samples of projection images with the segmentation mask corresponding to landmarks.

\begin{figure}[t!]
	\centering
	\includegraphics[width=0.7\linewidth]{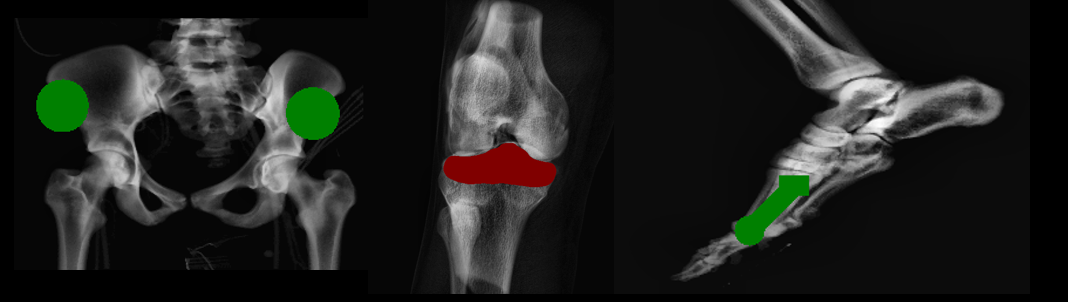}
	\caption{Projection images and the corresponding masks, for hip, tibial Plateau and foot metatarsal respectively}
	\label{fig:Extrem}
\end{figure}


For training, we utilized $60$, $40$, and $40$ 3D volumes of foot/ankle, knee, and hip, respectively. Prior to creating the projection images and masks, the 3D volumes were augmented via multiple 3D rotations and intensity adjustments. Additionally, we employed 10 volumes as a testing cohort for each of the anatomies. To assess incremental learning capabilities, we leveraged an in-house dataset of 3D head images and generated projection images using the same method.

\subsection{Architecture, Training and Testing}

The 2D nnUNet framework, as introduced by Isensee et al. in their work~\cite{isensee2018nnunetselfadaptingframeworkunetbased}, serves as the foundation for all our experiments, including the DoD-Net. To assess the effectiveness of our proposed label sharing approach, we compared it with three other methods 
\begin{enumerate}
\item Individual: Models trained separately for each task. These networks trained exclusively for a task type are expected to perform the best and hence will be  considered gold standard through our experiments,
\item Multichannel: All tasks merged into a single model,
\item DoD-Net: Incorporation of task-specific filters.
\end{enumerate}

We made modifications to the labels based on specific method employed. All ur models, for both use-cases, were trained for $100$ epochs using the Dice Similarity Coefficient (DSC) loss across all experiments. In the context of incremental learning within the label sharing framework, we fine-tuned the pretrained model (initially trained on the first set of tasks) for an additional $30$ epochs on the new task. Further, we performed combined training for 70 epochs by combining data from all tasks.

Our evaluation, both qualitative and quantitative, is elaborated in Section~\ref{sec:results}. We use label-wise and aggregate Dice scores to evaluate the performance of the segmentation models and Hausdorff distance for point and line landmarks respectively, for the extremities localization models.  

\section{Results}
\label{sec:results}

\subsection{2D segmentation task}
\begin{figure}[t!]
	\centering
	\includegraphics[width=\linewidth]{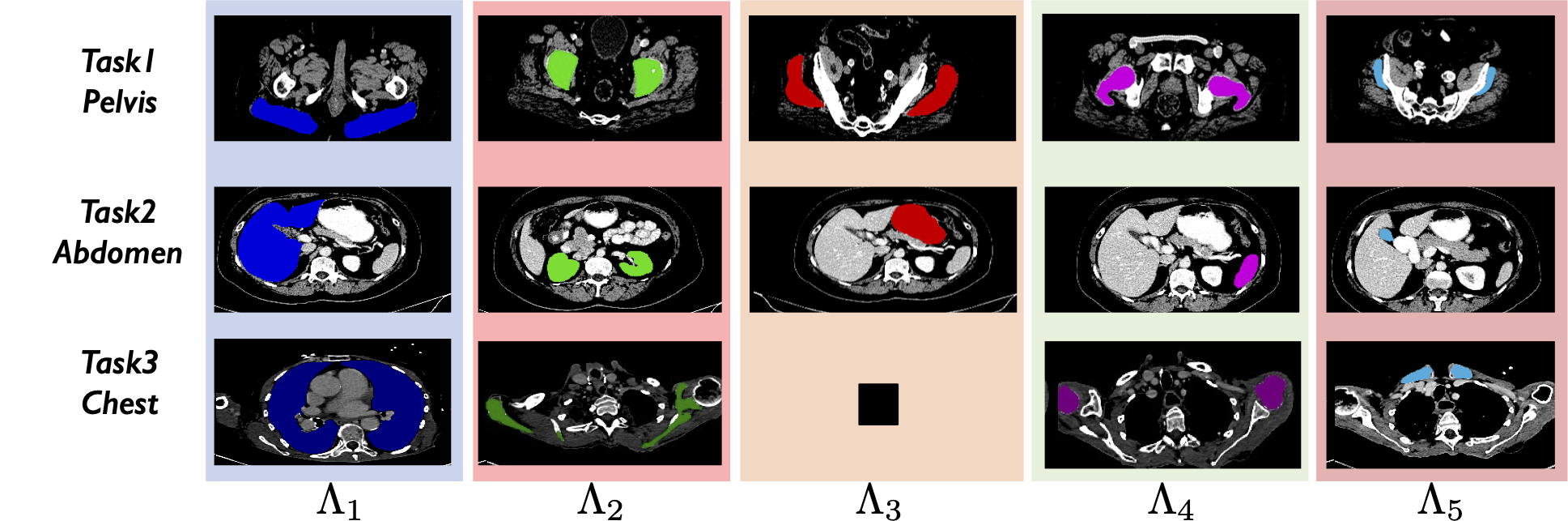}
	\caption{Illustration of segmentation performance across 14 anatomies (3 + 1 incremental tasks) on TotalSegmentator dataset}
	\label{fig:SegmentationQual}
\end{figure}

Figure~\ref{fig:SegmentationQual} provides a snapshot of the robust segmentation achieved for various organs across multiple tasks using the label sharing framework. As previously discussed, the network architecture for this set of $K$ tasks is designed with $5=\max(C_i)_{i=0}^K$ output channels (denoted as $C$). A visual examination of the segmentation results, spanning from large to small anatomical structures, demonstrates the performance robustness across all the label channels.

\begin{figure}
    \subfloat[Dice similarity coefficient compared for all methods with respect to total segmentator ground truth for all labels and GT provided in \cite{podobnik2023han} for Head \& Neck data]{\includegraphics[width=0.5\textwidth]{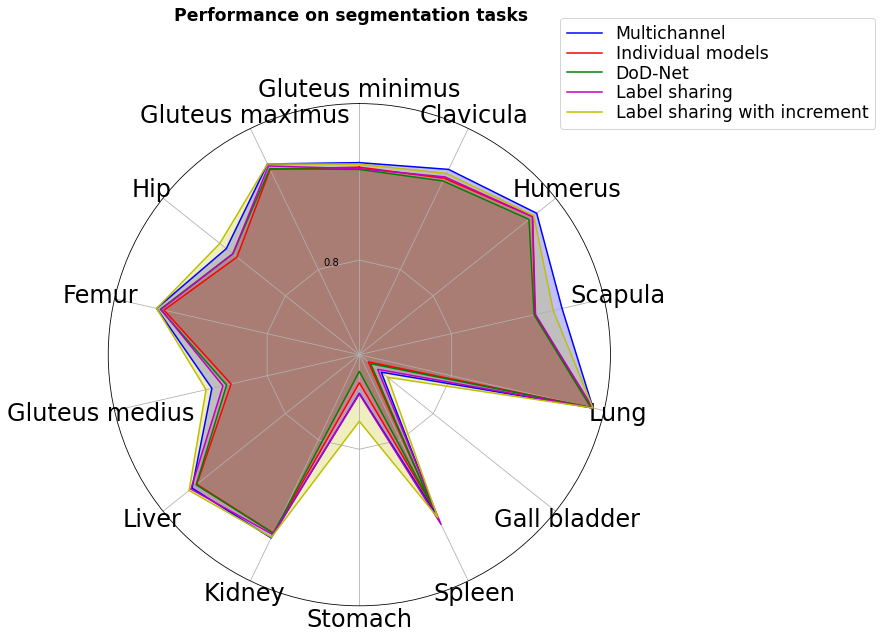}\label{subfig:DiceTotalSegmentator}}\hfill
    \subfloat[Comparitive evaluation of proposed Label Sharing method with individual task models for incremental tasks shows similar performance despite being parameter efficient ]{\includegraphics[width=0.37\textwidth]{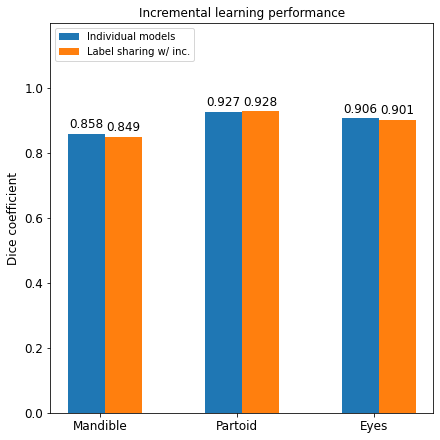}\label{fig:sub_b}}
    \caption{Dice comparison of the proposed Label Sharing methods with other methods shows no significant degradation compared to other models while offering incremental learning ability without modifying the architecture and a smaller parameter set}\label{fig:sub}
\end{figure}

\Cref{subfig:DiceTotalSegmentator} illustrates the Dice scores across different labels, along with a comparison to three other methods: individual models for each task, multi-channel output, and DoD-Net output. Naturally, the individually trained models exhibit the best performance. However, our proposed label sharing method consistently outperforms other approaches and closely approaches the performance of individually trained models in most cases. This highlights the efficacy of simplifying the network compared to more complex methods.
\begin{table}[h]
	\addtolength{\tabcolsep}{-1pt}
	\centering
	\caption{Dataset Group 1 -  Comparison of task-wise average Dice}
	\begin{tabular}{|c|c|c|c|c|}
		\hline
		& \textbf{Pelvic Region} & \textbf{Abdomen Region} & \textbf{Chest Region} & \textbf{Head Region}   \\ \hline
		\textbf{Multichannel}  &0.913 & 0.851 & 0.953 &  \\ \hline
		\textbf{DoD Model}  & 0.908 & 0.838 & 0.951 & \\ \hline
		\textbf{Label Sharing}  &  0.923 &	0.85 & 0.966 & \\ \hline
		\textbf{Label Sharing - IL} & 0.912 & 0.834 & 0.948 & 0.893\\ \hline
		\textbf{Individual} & 0.925 &	0.861 & 0.961 & 0.897 \\ \hline
	\end{tabular}
		\label{tab:tasks1eval}
\end{table}
Furthermore, incremental training for new tasks does not compromise the performance. The label sharing method maintains strong performance across both previous and new tasks, closely matching the best performance achieved by individual models. For a comprehensive overview, refer to Table~\ref{tab:tasks1eval}, which summarizes the average Dice Similarity Coefficient (DSC) performance for each task across the different methods.

%

\begin{figure}[t!]
    \subfloat[Normalized Hausdorff distance computed between results from each methods with respect to the ground truth generated by radiologist across 9 tasks. Smaller the value, better the performance]{\includegraphics[width=0.5\textwidth]{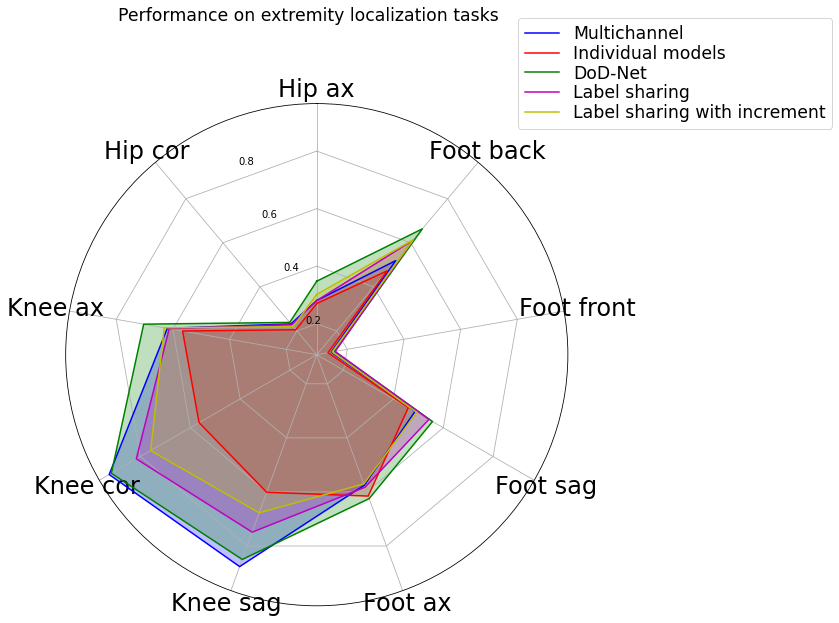}\label{fig:Haussdorff}}\hfill
    \subfloat[Comparison of Hausdorff between  the proposed label sharing method as compared to Individual models on incremental tasks shows improved performance even compared to individual models ]{\includegraphics[width=0.37\textwidth]{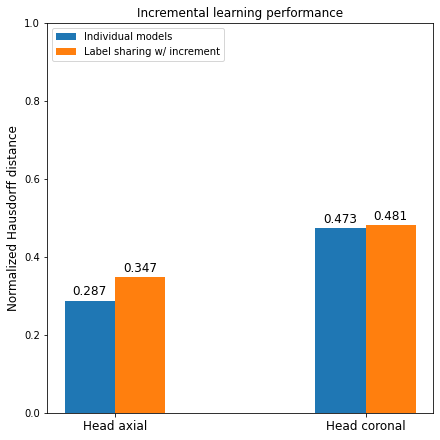}\label{fig:sub_b}}
    \caption{Comparison of Hausdorff distance between the different methods demonstrates higher effectiveness of the proposed method to other parallel methods other than individual models which are parameter heavy}\label{fig:sub}
\end{figure}


\subsection{Extremity localization task in 2D}
We observe a similar trend to the segmentation task as depicted in \cref{fig:Haussdorff}. In our analysis, we compute the Hausdorff distance  
between labels generated by each of the methods and the ground truth provided by the expert radiologist. To facilitate comparison, we normalize the absolute distance values with respect to the maximum Hausdorff distance.

A lower Hausdorff distance corresponds to better alignment with the ground truth. As expected, the individually trained models exhibit best results in most scenarios. However, the label sharing method consistently outperforms other approaches, except for coronal projection of knee. The lower performance for this task can be attributed to the fact that labels of long structures (tibia and femur) often get fragmented leading to confusion of label assignment in the label sharing framework. Apart from this particular scenario, the performance is better in comparison to the multi-channel and DoD-Net.  

Furthermore, the incrementally trained model demonstrates no adverse effects due to the newly added task. It performs comparably well to the individual models for both existing and new tasks.

%% file: sections/discussion.tex

This paper presents a novel framework for training a single model simultaneously on multiple segmentation tasks involving multiple disparate label . It has been shown by the experimental results that when labels across tasks are appropriately grouped, networks with small capacity are sufficient to achieve performance on par with models trained individually. The models trained with label sharing framework do not require any side information about the individual tasks. Interestingly, our experiments showed that the images corresponding to labels within a group need not exhibit semantic similarity for the model to successfully perform well on disparate datasets. In fact, as demonstrated, they need not even match with respect to absolute sizes. Additionally, the framework provides an elegant way to incorporate new datasets without changing the underlying model architecture or training procedure. 

An important line of work that remains to be done is automatic generation of shared labels, which may require new definitions of inter-task label similarity. Also, it remains to be studied how this framework extends to other imaging modalities, or even to multi-modal settings. 